# AraHopeCorpus: Annotation Guidelines and Dataset for Hope Speech in Arabic Social Media Crisis Discourse

**Esra'a Sharqawi, Wajdi Zaghouani**
Hamad Bin Khalifa University, Northwestern University in Qatar
Doha, Qatar
essh68661@hbku.edu.qa, wajdi.zaghouani@northwestern.edu

**Abstract**

Social media has become a crucial arena for shaping public narratives during armed conflicts, providing space for both harmful and constructive communication. While hate speech and misinformation have been widely studied, expressions that promote resilience, solidarity, and optimism remain underexplored, particularly in Arabic contexts. This paper introduces *AraHopeCorpus*, the first annotated dataset of Arabic hope speech collected from ten thousand YouTube comments related to the war on Gaza between 2023 and 2024. Using a detailed annotation framework, comments were classified into three categories: hope speech, no hope speech, and neutral or unclear discourse. The dataset shows that hopeful language dominates, accounting for more than sixty four percent of all comments. These expressions of hope appear mainly as religious encouragement, collective solidarity, and optimism for endurance and justice. No hope speech, representing about thirteen percent, reflects despair and disillusionment, while the rest of the comments contain neutral or mixed content. Inter-Annotator Agreement reached substantial levels (Cohen's Kappa equals 0.71), though dialectal variation, sarcasm, and implicit meaning posed annotation challenges. A comparative analysis between human annotators and ChatGPT revealed that large language models can support annotation but remain limited in handling dialectal and culturally embedded expressions. *AraHopeCorpus* will be released for research purposes under an open and non commercial license. It provides a valuable resource for studying constructive digital discourse, enabling further research on hope speech detection, crisis communication, and resilience in Arabic social media.

## 1. Introduction

Social media platforms such as YouTube can function as archives of human suffering, testimonies, and global solidarity. The Gaza genocide[1] (2023 to 2024), which claimed over forty eight thousand Palestinian lives and displaced millions, represents a paradigmatic case where digital discourse reflects both trauma and resilience. Existing research overwhelmingly highlights the toxicity of online spaces during crises, focusing on hate speech, propaganda, and misinformation. However, constructive forms of discourse have received far less attention. Hope speech, defined as messages that inspire perseverance, healing, and unity, offers a critical counter narrative to despair. While studies have explored hope speech in South Asian and European languages (Chakravarthi, 2022)(Arif et al., 2024), no systematic effort has been made to examine its presence in Arabic digital conflict discourse.

This paper fills that gap by introducing the first large scale annotated dataset of Arabic hope speech from YouTube comments during the Gaza genocide. By analyzing ten thousand comments, we show that hope oriented narratives dominate digital spaces, functioning as linguistic markers of resilience, faith, and community. The dataset is designed to serve multiple purposes: training computational models for hope speech detection, understanding digital resilience mechanisms, and providing empirical evidence of constructive discourse in crisis contexts.

To guide the study, we formulated the following research questions:

- RQ1: What are the dominant themes of hope speech in Arabic language YouTube comments on the Gaza genocide?
- RQ2: How does hope speech compare to other forms of discourse (No Hope Speech and Neutral or Unclear Speech) in the dataset?
- RQ3: How efficient is ChatGPT in annotating Arabic hope speech during crisis, and what are its limitations compared to human annotators?
- RQ4: How does the Inter Annotator Agreement (IAA) reflect the complexity of hope speech annotation, and what factors contribute to annotation disagreements?
- RQ5: What are the most frequently used words and phrases in hope speech, and what do they reveal about the nature of digital resilience and solidarity?

By answering these questions, the study contributes a novel annotated dataset for Arabic NLP

[1] The United Nations Commission of Inquiry on the Occupied Palestinian Territory qualified Israel's war on Gaza as an act of genocide. See: UN OHCHR, 2025 Press Release.

and a deeper understanding of digital resilience narratives during times of war.

## 2. Related work

Research relevant to this study spans several overlapping domains: computational approaches to social media in conflict, natural language processing (NLP) in Arabic, and the emerging field of hope speech analysis

### 2.1. Social Media and Conflict

Social media has become a central arena in which conflicts are documented, debated, and framed. Previous studies have examined how online spaces amplify violence through hate speech (Fortuna and Nunes, 2018; Schmidt and Wiegand, 2017) or spread misinformation and propaganda (Asimovic and Esberg, 2022) . In the Middle East, researchers have documented how Twitter, Facebook, and YouTube function as archives of collective trauma as well as resistance (Khamis, 2022). For Gaza specifically, studies highlight how digital platforms create counter-narratives to mainstream media biases, amplifying Palestinian suffering and resilience (Liyih et al., 2024). Recent work has also examined emotional and trauma-related discourse in Arabic social media during crisis events. For instance, Shurafa and Zaghouani (2024) analyze large-scale YouTube comments to study emotional responses and coping mechanisms expressed in Arabic online communities during traumatic situations.

However, most computational work on conflict discourse has overwhelmingly focused on harmful content detection, with very little attention to constructive or hopeful discourse

### 2.2. NLP and Arabic Social Media

Arabic NLP poses significant challenges due to dialectal variation, code-switching with English or French, and non-standard orthography(Habash, 2010). Over the past decade, a number of resources have been developed to address abusive or hateful content in Arabic. These include the Arabic Offensive Language Dataset (AOLD) (Mubarak et al., 2020)(, the So Hateful! Dataset (Zaghouani et al., 2024a), and the ADHAR Corpus ((Charfi et al., 2024)for hate speech in Arabic dialects.

Several large-scale Arabic social media corpora have also been developed to study linguistic variation, demographic attributes, and discourse patterns in online communication (Zaghouani and Charfi, 2018). In addition, recent work has introduced new annotated datasets targeting abusive and hateful language in Arabic social media, further advancing research on harmful online discourse (Zaghouani et al., 2024b).

Such datasets have advanced the development of machine learning and deep learning models for content moderation in Arabic. Nevertheless, the focus has almost exclusively been on toxic or harmful speech, neglecting other dimensions of digital discourse such as resilience, humor, or hope. This leaves a significant gap in resources for constructive discourse modeling in Arabic.

### 2.3. Hope Speech Research

Hope speech as a computational category is relatively new. (Chakravarthi, 2022) proposed one of the first annotation schemes for identifying hope-oriented messages in English and Tamil, defining it as language that inspires optimism and perseverance. (Arif et al., 2024) expanded this with PolyHope, a multilingual dataset covering English, Tamil, Malayalam, and Hindi, confirming that hope speech can be systematically identified across languages. More recently, research has begun to explore multilingual datasets that combine emotion and hope-oriented discourse in social media communication. For example, Zaghouani and Biswas (2025) introduce EmoHopeSpeech, a bilingual dataset designed to analyze the relationship between emotional expression and hope speech across languages. These efforts further demonstrate the growing interest in modeling constructive and resilience-oriented discourse in NLP. These studies showed the potential of hope speech detection models to contribute to peacebuilding, mental health interventions, and civic discourse. Yet, Arabic remains absent from this line of research. The only partial references to constructive Arabic discourse appear in studies of religious invocations in crisis communication (Al-Saidi, 2021)( or the use of solidarity slogans in protest contexts (Abdelali et al., 2020). To date, no large-scale Arabic hope speech dataset has been created, particularly in a conflict setting such as Gaza.

### 2.4. Annotation and Human–AI Comparison

Annotation of affective discourse is challenging due to subjectivity, sarcasm, and dialect-specific meanings. Previous work has compared human annotation with LLM-assisted annotation (Bang et al., 2023), noting that while models like ChatGPT can speed up annotation, they struggle with nuanced categories and under-represented dialects. In Arabic, these limitations are magnified due to morphological richness and lack of training data. Our study therefore contributes in two ways:

1. By providing the first annotated Arabic hope

speech dataset, filling a gap in resources for constructive discourse.

2. By evaluating both human annotation reliability (IAA) and the efficiency and limitations of ChatGPT annotation for Arabic, adding to current debates about the role of LLMs in data creation.

# 3. Methodology

## 3.1. Data Sources and Selection

The dataset was built from 23 carefully selected YouTube videos covering the Gaza genocide (2023–2024). These videos were chosen to represent a diverse mix of sources:

- Mainstream Arabic news channels (Al Jazeera, BBC Arabic, Syria TV)
- Independent Palestinian journalists such as Wael Dahdouh and Saleh Al-Ja'frawi, whose on-the-ground reporting gained wide traction
- Activist accounts documenting atrocities, civilian experiences, and humanitarian appeals

Selection criteria included:

1. Relevance: Videos had to explicitly reference Gaza events between October 2023 and November 2024, during the Gaza Genocide.
2. Engagement: Each video contained at least 800 comments.
3. Credibility: Sources were cross-checked against known disinformation channels.

The final video selection included:

- 20 news reports videos from established media outlets
- 3 videos from independent journalists and influencers

## 3.2. Data Collection Process

Using the Communalytic platform, we extracted over 82,000 comments, preserving metadata such as user ID (anonymized), timestamps, and thread structures. The data collection occurred in April 2024. Cleaning involved: (1) Removing duplicates and spam (identified through exact match and fuzzy string matching); (2) Filtering out non-Arabic text, emoji-only comments, and URLs ; (3) Normalizing Arabic orthography (e.g., removing diacritics, unifying i̲ , i.J) and (4) Removing bot-generated comments (identified by posting patterns and repetitive content)

After filtering, a balanced 10,000-comment dataset was selected for annotation, ensuring representation across all video types and time periods. The sampling strategy prioritized: (1) Proportional representation from each video source type; (2) Temporal distribution across conflict phases and (3) Variety in comment length (short vs. extended discourse)

# 4. Annotation Guidelines

This study employed a systematic framework for classifying user-generated comments into three primary categories: Hope Speech (HS), No Hope Speech (NHS), and Neutral/Unclear (NU), designed to capture nuanced expressions of sentiment and attitude within social media discourse. Table 1 summarizes the subcategories for each class.

| Category | Subcategories |
|---|---|
| **HS** | Support/Encouragement; Unity/Peace; Optimism for the Future; Religious/Spiritual Encouragement; Calls for Justice; Strength/Endurance Recognition; Compassion/Empathy; Collective Action for Change |
| **NHS** | Despair/Frustration; Criticism without Hope; Pessimism; Critique of the Cause Itself; Distrust/Disillusionment with Leadership |
| **NU** | General Observations; Mixed Messages; Ambiguous Statements; Philosophical or Reflective; Condemnation of Violence; Critique of Oppressors' Actions; Labeling the Oppressors; Irrelevant |

Table 1: Annotation scheme: categories and subcategories.

Hope Speech encompasses comments that actively promote optimism, resilience, or collective solidarity in response to the ongoing crisis. Within this category, religious and spiritual encouragement represents a particularly salient subcategory in the Arabic context, where faith-based invocations function as anchors of psychological endurance.

Expressions of support and encouragement, calls for justice, and recognition of collective strength further reflect the diverse ways in which hope manifests linguistically, ranging from direct motivational appeals to implicit moral endorsement of resistance and perseverance.

No Hope Speech captures expressions marked by despair, disillusionment, or the absence of con-structive outlook. This category is distinguished from mere negativity by its orientation toward hopelessness rather than critique with intent. Criticism without hope, for instance, differs from calls for justice in that it offers no forward-looking dimension.

Distrust and disillusionment with leadership and pessimism similarly reflect fatigue and loss of faith in collective agency, making this category particularly sensitive to context and tone. The Neutral/Unclear category accommodates comments that resist binary classification, including ambiguous or sarcastic statements, philosophical reflections, and factual observations about events. This category proved especially challenging during annotation, as sarcasm and implicit meaning in dialectal Arabic frequently obscured intent.

The classification process therefore considered both explicit content and implicit meaning, acknowledging that hope or its absence may be conveyed through culturally specific linguistic and rhetorical frameworks. Guidelines were updated iteratively throughout the annotation process as new edge cases emerged.

| Category | Example |
| --- | --- |
| Hope Speech (HS) | [illegible] (Victory is near, God willing) |
| No Hope Speech (NHS) | [illegible] (We have no hope in this unjust world). |
| Neutral/Unclear (NU) | [illegible] (The bombing happened at night) |

Table 2: Example of comments from the three categories

The classification process considered both explicit content and implicit meaning, acknowledging that hope or its absence might be expressed through various linguistic and cultural frameworks. The annotation guidelines were updated iteratively as new, uncategorized comments emerged during the coding process.

## 4.1. Annotation Team and Training

Three trained annotators, all native Arabic speakers fluent in multiple dialects (Levantine, Egyptian, Gulf), conducted the labeling. All annotators held graduate degrees and had prior experience with Arabic text annotation. The process involved:

1. Pilot phase: 300 comments to calibrate judgments and identify ambiguities
2. Guideline revision: After pilot phase, guidelines were refined based on annotator feedback, particularly around religious expressions and sarcasm
3. Full annotation: Conducted over six weeks with weekly calibration meetings Double-coding: 25% of the dataset (2,500 comments) was independently annotated by all three annotators for reliability measurement

Annotators used a custom web-based annotation interface that displayed comments with video context (title, description) to aid interpretation.

## 4.2. Inter-Annotator Agreement (IAA)

**Evaluation set.** To assess reliability, we use the portion of the corpus that was annotated independently by all three trained annotators (25% of the data, 2,500 comments). Labels follow our three-way scheme {Hope, No Hope, Neutral} described in §4 (*Annotation Guidelines*).[2] This subset was produced during the planned double-coding phase with weekly calibration meetings. ChatGPT is not included in human agreement statistics; its outputs are compared to the human consensus in §4.3.

**Metrics.** We report (i) raw percentage agreement $A$ and (ii) pairwise Cohen's $\kappa$ between each pair of human annotators, followed by a macro-average across the three pairs. Raw agreement is

$$A = \frac{1}{N} \sum_{i=1}^{N} 1\{y_i^{(a)} = y_i^{(b)}\},$$

where $N$ is the number of items and $y_i^{(a)}$ and $y_i^{(b)}$ are the labels assigned by annotators $a$ and $b$. Cohen's kappa is

$$\kappa = \frac{p_o - p_e}{1 - p_e},$$

with $p_o$ the observed agreement (equal to $A$) and $p_e$ the chance agreement computed from the empirical label distributions of the two annotators. Because classes are imbalanced, we additionally report category-wise agreement rates against the primary analysis annotator (Table 7), which helps interpret $\kappa$ under skewed label distributions.

**Results and interpretation.** Across the triple-coded subset, human agreement is *substantial*; the corpus-level macro-average Cohen's $\kappa$ is $0.71$. This indicates consistent application of the scheme despite dialectal variation, sarcasm, and implicit religious expressions that make the task challenging. Similar annotation challenges have been observed in other Arabic social media datasets focusing on implicit meanings such as irony and figurative language (Abbes et al., 2020). Category-wise rates in Table 7 further show higher alignment on explicit Hope cues and more divergence on No Hope versus Neutral, which is expected for culturally nuanced, context-dependent content. ChatGPT statistics are reported separately in §4.3 to distinguish human reliability from model comparison.

[2] See §4 for category definitions and examples.

### 4.3. ChatGPT-Based Annotation Experiment

We conducted a parallel annotation experiment using ChatGPT-4, comments were annotated using zero-shot and few-shot prompting strategies:

- Zero-shot prompt: Provided category definitions and asked for classification
- Few-shot prompt: Included 10 examples per category before requesting classification

ChatGPT's annotations were compared against the gold standard (human consensus labels) using:

- Accuracy, Precision, Recall, and F1-score per category
- Error analysis to identify systematic failures
- Qualitative review of misclassified examples

**Advanced Prompting Strategies.** In addition to the zero-shot and few-shot configurations reported in this study, we conducted preliminary pilot experiments with more advanced prompting strategies, including role-based prompting and chain-of-thought reasoning prompts. These approaches were designed to encourage the model to explicitly reason about contextual and cultural cues within Arabic comments. However, initial observations suggested that these strategies did not substantially improve classification performance, particularly for dialectal expressions and implicit religious references that characterize hope speech in Arabic discourse. Because the primary objective of this paper is to introduce and validate the AraHopeCorpus dataset and annotation framework, a systematic evaluation of advanced prompting and fine-tuning strategies is left for future work.

## 5. Results

As shown in Table 2, hope speech predominates in the dataset, constituting 64.3% (6,430) of all comments. This indicates a strong tendency toward positive and resilient discourse despite the challenging circumstances. No hope speech accounted for 13.5% (1,350) of comments, while neutral or unclear speech comprised 22.2% (2,220) of the dataset. This distribution suggests that most YouTube commenters engaged with content about Gaza through expressions of support, encouragement, and faith rather than expressions of pessimism or disengagement.

| Category | Count | Percentage |
|---|---|---|
| Hope Speech | 6,430 | 64.3% |
| No Hope Speech | 1,350 | 13.5% |
| Neutral/Unclear | 2,220 | 22.2% |
| **Total** | **10,000** | **100%** |

Table 3: Distribution of speech categories in Gaza-related YouTube comments

### 5.1. Analysis of Hope Speech Distribution

| Hope Speech Subcategory | Count | Percentage |
|---|---|---|
| Religious / Spiritual Encouragement | 3,246 | 50.48% |
| Support / Encouragement | 2,166 | 33.68% |
| Acknowledgment of Strength / Endurance | 723 | 11.24% |
| Collective Action for Change | 106 | 1.65% |
| Compassion and Empathy | 81 | 1.26% |
| Optimism for Future | 56 | 0.87% |
| Calls for Justice | 28 | 0.44% |
| Unity / Peace | 24 | 0.37% |
| **Total** | **6,430** | **100%** |

Table 4: Distribution of hope speech subcategories

The hope speech category, which dominated the dataset, exhibited diverse manifestations of positive discourse across eight subcategories. Religious and spiritual encouragement emerged as the most prevalent form (Table 3), comprising 50.48% (3,246) of all hope speech comments. This finding highlights the significant role of faith-based expressions as a coping mechanism and source of resilience within the discourse surrounding Gaza Genocide. Support and encouragement constituted the second largest sub-category, accounting for 33.68% (2,166) of hope speech comments. These expressions demonstrated solidarity, validation, and motivation toward those affected by the genocide. Acknowledgment of strength and endurance appeared in 11.24% (723) comments, rec-ognizing the perseverance and resilience demonstrated amid adversity.

### 5.2. Analysis of no hope speech distribution

| No Hope Speech Subcategory | Count | Percentage |
|---|---|---|
| Criticism Without Hope | 1,005 | 74.4% |
| Distrust / Disillusionment with Leadership | 184 | 13.6% |
| Despair / Frustration | 128 | 9.5% |
| Critique of the Cause Itself | 33 | 2.4% |
| **Total** | **1,350** | **100%** |

Table 5: Distribution of no hope speech subcategories

As shown in Table 4, Criticism without hope emerged as the dominant subcategory, accounting for 74.4% (1,005) of all no hope speech comments. These expressions reflected frustration and critical assessments of the situation without offering constructive alternatives or expressions of optimism. Distrust and disillusionment with leadership constituted the second most common form of no hope speech, present in 13.6% (184) of comments within this category. This finding indicates significant skepticism toward political figures and governance structures associated with the conflict. Expressions of despair and frustration appeared in 9.5% (128) of no hope speech comments, highlighting the emotional toll of the ongoing situation. Critique of the cause itself (2.4%, 33 comments) was relatively rare in the dataset.

### 5.3. Analysis of neutral/unclear speech distribution

As shown in Table 5, the neutral/unclear speech category made up 22.2% of all comments and included varied expressions that resisted clear classification as hopeful or despairing. Within this group, general observations were most common (31.8%, 706 comments), followed by mixed messages combining hope and despair (28.4%, 631), and ambiguous statements with unclear sentiment (12.1%, 269). Other subcategories included irrelevant comments (11.6%, 257), labeling of oppressors (9.0%, 200), philosophical reflections (4.5%, 99), critiques of oppressors' actions (1.4%, 30), and condemnations of violence (1.3%, 28). These figures indicate that neutral discourse was dominated by factual or mixed expressions, with moral or reflective content appearing less frequently.

| Neutral/Unclear Speech Subcategory | Count | Percentage |
|---|---|---|
| General Observations | 706 | 31.8% |
| Mixed Messages | 631 | 28.4% |
| Ambiguous Statements | 269 | 12.1% |
| Irrelevant | 257 | 11.6% |
| Labeling the Oppressors | 200 | 9.0% |
| Philosophical or Reflective | 99 | 4.5% |
| Critique of Oppressors' Actions | 30 | 1.4% |
| Condemnation of Violence | 28 | 1.3% |
| **Total** | **2,220** | **100%** |

Table 6: Distribution of neutral/unclear speech subcategories

### 5.4. Temporal Trends in Hope Speech

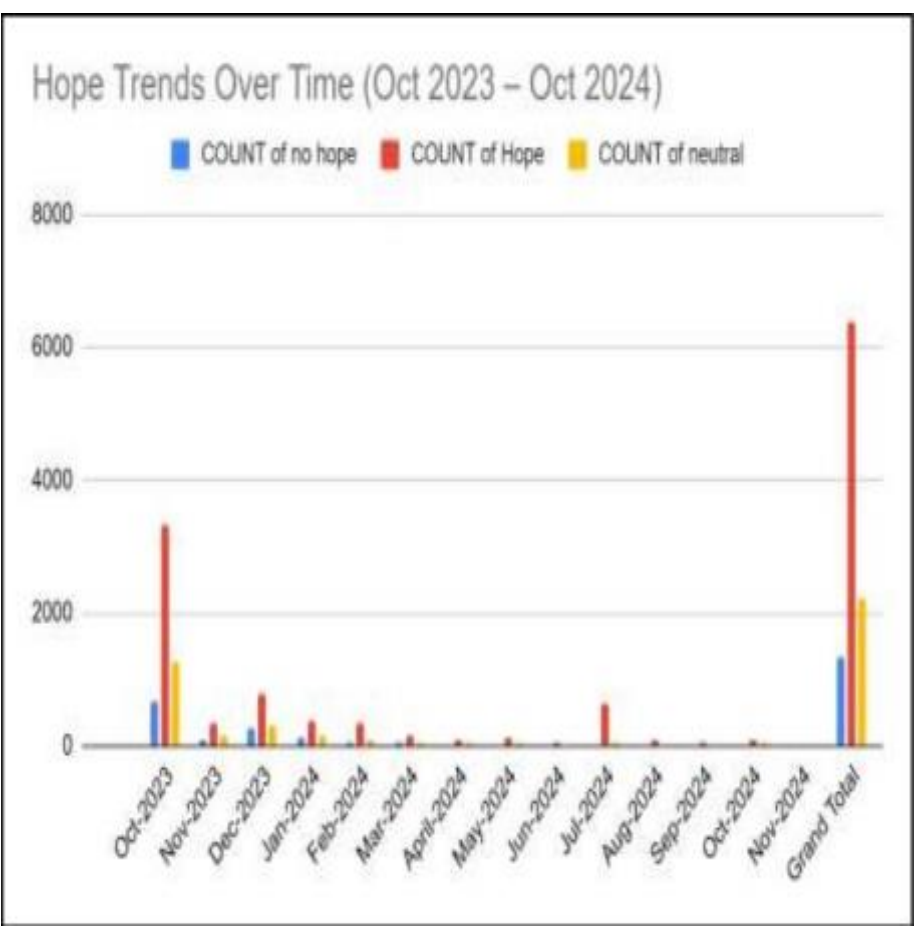

Figure 1: Temporal Distribution of Categories

Figure 1 "Temporal Distribution of Categories" illustrates fluctuations in hope-related sentiments across 10,000 YouTube comments on the Gaza genocide. Hope speech, representing 64.3% of all comments, remained consistently visible but peaked in October 2023 and October 2024, likely tied to major conflict events and driven by religious or spiritual encouragement (50.5% of hope speech). In contrast, no-hope speech (13.5%) rose sharply in October 2024, reflecting increased frustration, as the subcategory "criticism without hope" dominated (74.4%). Neutral speech (22.2%) also increased toward the end of the period, showing a parallel rise with hope and no-hope expressions and underscoring the evolving and complex nature of public sentiment over time.

## 5.5. Words frequency of hope speech vs. no hope speech

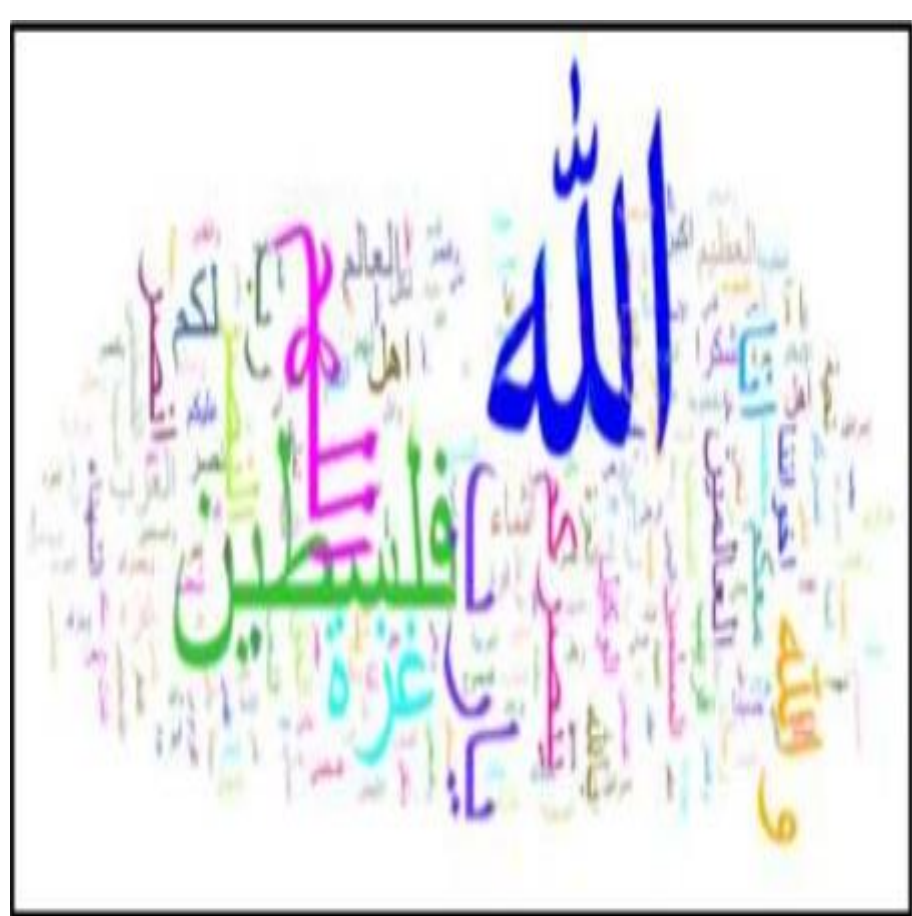

Figure 2: Word cloud of hope speech frequent words(Voyant tools)

The hope speech category, which formed the majority of the dataset, was dominated by religious and spiritual encouragement (50.48%), reflected in frequent references to -.lJJ (Allah), •lJJ (O Allah), and _ _l, (O Lord), illustrating faith as a key source of resilience. This was followed by support and encouragement (33.68%), where mentions of u· ,.k - l · (Palestine) and -0 · „· (Gaza) expressed steadfast solidarity. Acknowledgment of strength and endurance (11.24%) appeared through the use of _.,.. .· JJ (victory), symbolizing enduring optimism. By contrast, activism-oriented expressions—calls for justice (0.44%) and collective action (1.65%)—were limited, suggesting that hope was largely articulated through spiritual and emotional solidarity rather than direct political mobilization.

The no-hope speech category, comprising 13.5% of the dataset, was primarily defined by criticism without hope (74.4%), reflected in frequent mentions of _ _ - JJ (Arabs), ilc,. (rulers), and tJl-JJ (the world), expressing frustration with Arab leadership and global inaction. Distrust and disillusionment with leadership (13.6%) appeared in references to u· ,-4l_-_ lJ (Muslims) and -_, _-JJ (Arab identit y/language), indicating disap pointment in collective solidarity. Expressions of despair and frustration (9.5%) featured terms like -.. .· ,lrD,.JJ (Zionists) and ..rl.."" (Hamas), reflecting polarized views on the ongoing genocide. Overall, while hope speech centered on faith and perseverance, no-hope speech conveyed deep skepticism, disillusionment, and frustration toward leadership and the international community.

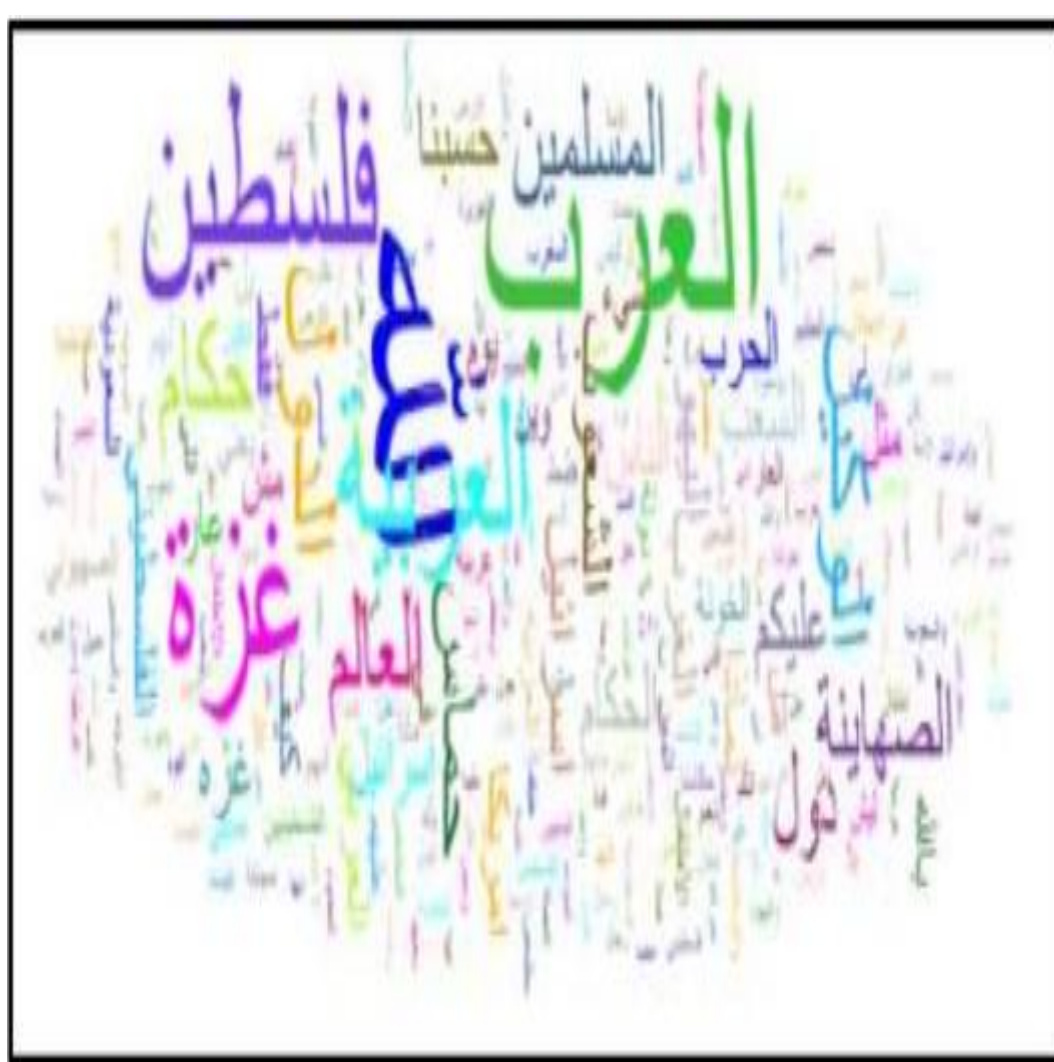

Figure 3: Word cloud of no hope speech frequent words(Voyant tools)

## 5.6. Inter Annotation Agreement analysis

| Category | A2 (%) | A3 (%) | A4 (%) | A5 (%) |
|---|---|---|---|---|
| Neutral | 35.23 | 62.50 | 63.64 | 80.68 |
| Hope | 94.06 | 83.75 | 88.44 | 18.13 |
| No Hope | 71.43 | 45.45 | 54.55 | 0.00 |

Table 7: Annotators' classification results in percentage. Annotator 5 corresponds to the ChatGPT model.

All annotators followed standardized guidelines for the three categories: hope, no hope, and neutral and their results were compared with the primary annotator to assess consistency. Human annotators showed strong agreement for hope speech (Annotator 2: 94.06%, Annotator 4: 88.44%, Annotator 3: 83.75%) and moderate alignment for no-hope speech (71.43%, 45.45%, and 54.55%, respectively). Agreement for neutral content ranged from 35.23% to 63.64%.

The automated annotator (ChatGPT) achieved high accuracy for neutral content (80.68%) but performed poorly for hope (18.13%) and no hope (0.00%). These results highlight the difficulty of detecting nuanced emotional and culturally embedded expressions through automation. Methodological issues included task comprehension, overreliance on lexical cues, and challenges in interpreting dialectal or context-dependent meanings, un-derscoring the continued importance of human expertise in sentiment annotation.

Beyond these quantitative findings, the following discussion highlights key linguistic and cultural patterns emerging from the annotations, along with insights into the performance of human and AI annotators.

### 5.7. Discussion of Results

The analysis of *AraHopeCorpus* reveals that hope speech dominates Arabic digital discourse during the war on Gaza, representing about sixty-four percent of all comments. These messages frequently combined religious encouragement, moral support, and collective endurance, showing how faith-based optimism functions as an anchor of social and emotional resilience. In contrast, no-hope speech, accounting for roughly thirteen percent, conveyed frustration, fatigue, and disappointment toward leadership or international responses, while neutral and mixed comments reflected dialogic reflection rather than polarization.

From a linguistic perspective, hope speech relied heavily on metaphor, repetition, and religious expressions that linked perseverance to divine justice. These findings demonstrate that positive emotional communication persists even amid large-scale crisis and that Arabic social media provides a distinct space for communal coping and moral solidarity. The results also confirm that hope speech is not monolithic but context-sensitive, varying across dialects and thematic triggers.

The comparison between human annotators and ChatGPT underscores both the potential and the limits of AI-assisted annotation. While the model effectively detected explicit expressions of optimism, it frequently misclassified implicit or dialectal hope and failed to capture spiritual nuance. Human annotators achieved substantial agreement ($\kappa = 0.71$), indicating consistent application of the annotation scheme and highlighting the need for culturally informed human judgment. These insights emphasize that hybrid annotation pipelines—combining AI efficiency with expert review—offer a practical path for scalable, ethical corpus development.

From a computational perspective, these findings highlight the complexity of modeling constructive discourse categories that depend heavily on pragmatic and cultural cues. Unlike many sentiment classification tasks that rely on surface polarity markers, hope speech in Arabic often emerges through indirect linguistic signals such as religious invocations, metaphorical expressions, and references to collective identity. These characteristics suggest that lexical features alone are insufficient for reliable automatic detection and that future modeling approaches should incorporate contextual embeddings and discourse-level representations capable of capturing implicit meaning. Transformer-based architectures pretrained on Arabic social media text may provide a suitable foundation, particularly when combined with fine-tuning strategies that account for dialectal variation and pragmatic context. In addition, hybrid approaches integrating semantic representations with culturally grounded lexical resources could further improve the detection of nuanced hope-oriented expressions.

Overall, the discussion of results positions *AraHopeCorpus* as a resource that advances Arabic NLP beyond toxicity detection toward the modeling of constructive, empathetic, and resilience-oriented discourse. The dataset's findings illustrate how language and technology intersect in representing collective endurance and how future models can be trained to recognize not only harm but also hope. These findings also suggest that improving automated detection of hope-oriented discourse will likely require domain-specific training data and culturally informed modeling strategies. Future research may therefore explore fine-tuning large language models on dialectal Arabic corpora and incorporating contextual features such as discourse structure, pragmatic markers, and religious expressions that frequently signal hope in Arabic online communication.

### 5.8. Data Availability

AraHopeCorpus will be made available for research purposes via the following link[3]. The distributed version includes comment texts, category and subcategory labels, and video-level metadata (source type and collection period). Access is subject to a data-sharing agreement requiring (a) noncommercial academic use, (b) institutional ethics approval or equivalent waiver, and (c) a prohibition on re-identification of individuals. All resources are released under a Creative Commons Attribution NonCommercial ShareAlike 4.0 (CC BY-NC-SA 4.0) License, consistent with the ethical considerations discussed in the ethics statement section. Researchers are advised to consult YouTube's current Terms of Service regarding reuse of derived social-media data.

## 6. Implications and Broader Societal Impact

*AraHopeCorpus* exposes multiple challenges that remain central to advancing Arabic NLP and computational social science. Linguistically, the annotation of hope-oriented discourse in Arabic is complicated by dialectal variation, code-switching, and figurative or faith-based expressions. Models often misclassify such nuanced optimism as neutral or irrelevant, revealing a persistent gap in contextual understanding. Addressing these issues requires context-aware, human-in-the-loop annotation frameworks and training data that represent regional diversity, spiritual registers, and pragmatic subtleties. Collaboration between computational linguists, social scientists, and cultural experts is essential to reduce misinterpretation and ensure equitable representation of Arabic voices.

[3]https://tinyurl.com/4ke5jwyw

From a technical standpoint, this work highlights the need for annotation guidelines and model architectures that move beyond lexical features to incorporate semantics, discourse markers, and cultural pragmatics. Future work could employ multimodal approaches integrating textual, visual, and acoustic signals to better capture affective and moral dimensions of communication in crisis settings. Methodologically, the integration of large language models such as ChatGPT has demonstrated efficiency but also clear limitations: AI systems tend to underperform on dialectal or context-dependent content, underscoring the ongoing importance of expert human oversight.

Ethically, the dataset raises important considerations about privacy, consent, and data governance. Although all identifiable user information was removed and data were collected from publicly available sources, researchers are encouraged to apply this resource responsibly and to interpret findings within cultural and humanitarian contexts. Misuse scenarios such as sentiment surveillance, political profiling, or reductionist interpretations must be actively guarded against through transparent documentation, non-commercial licensing, and adherence to responsible-use guidelines. The corpus aligns with current recommendations in ethical NLP for transparency, explainability, and minimizing algorithmic harm.

At the sociopolitical level, the coexistence of solidarity and disillusionment within the dataset illustrates how social media simultaneously unites and divides communities during crises. Hope and despair frequently appear intertwined, reflecting complex attitudes toward institutions, leadership, and collective resilience. The computational study of such discourse contributes to digital humanities and social science by providing empirical insight into how emotional communication functions as a civic resource during collective hardship. This has potential applications for humanitarian communication, online community health, and digital peacebuilding.

Beyond research, the broader societal value of *AraHopeCorpus* lies in its contribution to inclusive and socially aware AI. By modeling empathy, compassion, and moral endurance, the dataset supports the development of systems that detect and amplify constructive communication. It offers the research community a benchmark for studying non-toxic, prosocial language in Arabic and a platform for investigating resilience, identity, and solidarity online. The resource can inform educational initiatives, policy-oriented language monitoring, and cross-cultural analysis of emotional discourse. In future work, expanding the dataset across platforms and modalities will allow for longitudinal tracking of hope and resilience narratives. Combining computational approaches with qualitative discourse and ethnographic analysis could reveal how digital expressions of hope evolve into collective action and offline recovery. In this sense, *AraHopeCorpus* stands not only as a linguistic dataset but also as a cultural and societal record—documenting how communities use language and technology to sustain empathy and meaning amid crisis.

# 7. Conclusion

This paper introduced *AraHopeCorpus*, the first large-scale annotated dataset of Arabic hope speech collected from YouTube comments during the war on Gaza. Through the annotation of ten thousand comments, we demonstrated that expressions of hope, faith, and collective solidarity play a crucial role in shaping digital resilience under crisis. The analysis showed that hope-oriented discourse significantly outweighed despairing or neutral messages, reflecting how social media can also serve as a site of moral support and communal endurance in times of suffering.

The proposed annotation scheme, inter-annotator evaluation, and comparative use of ChatGPT highlight both the potential and the limitations of human–AI collaboration in identifying culturally grounded expressions of hope. Beyond providing a benchmark for Arabic NLP, this dataset encourages interdisciplinary exploration of how language fosters coping and psychological resistance in the face of collective trauma.

Future research can build on this work by extending hope speech analysis to other dialects, platforms, and multimodal contexts, as well as by developing models capable of detecting constructive and empathy-driven discourse. *AraHopeCorpus* will be released for research purposes under an open and non-commercial license to promote transparency, inclusivity, and responsible development of language technologies that amplify voices of compassion and resilience.

# 8. Limitations

While *AraHopeCorpus* represents a substantial contribution to Arabic NLP and computational social science, several limitations should be acknowl-edged. These constraints pertain to data scope and coverage, annotation and methodological design, and generalizability.

### 8.1. Dataset Scope and Coverage

The dataset is derived exclusively from YouTube comments collected during a specific six-month period of the war on Gaza. Consequently, it reflects the linguistic and emotional dynamics of a single platform and timeframe rather than the full diversity of Arabic online discourse. Other social networks such as Twitter, Facebook, and TikTok exhibit different interaction norms and user demographics that could yield additional perspectives. Although the selected videos encompass diverse content sources, they still represent a subset of Gaza-related material, which may not capture the full range of public sentiment or smaller community voices. Moreover, while multiple Arabic dialects are represented, Levantine and Egyptian varieties dominate due to demographic and geographic factors, leaving Gulf, Maghrebi, and Sudanese dialects comparatively underrepresented. The dataset therefore provides a valuable but partial view of Arabic digital expression.

### 8.2. Annotation and Methodological Constraints

Hope speech is a highly context-dependent and subjective phenomenon. Annotators sometimes differed in interpreting whether a comment conveyed hope, sarcasm, or resignation, particularly when the expression relied on religious references, irony, or implicit emotion. Despite a substantial inter-annotator agreement score ($\kappa = 0.71$), these ambiguities reveal the inherent difficulty of categorizing affective discourse in multilingual and culturally loaded contexts. The annotation process was supported by ChatGPT, whose performance highlighted both the potential and limitations of large language models in handling dialectal and spiritually nuanced text. While AI assistance increased efficiency, the model occasionally failed to identify implicit optimism or contextually bound meaning. Future annotation pipelines should combine automated pre-labeling with expert human validation to improve both scalability and interpretive depth.

Contextual understanding was also limited by the available metadata. Annotators viewed comments with minimal video context and without detailed information on user background, reducing the ability to assess sociolinguistic variation or conversational flow. A more comprehensive approach incorporating full thread structures, multimodal inputs, or user-provided metadata would enrich future analyses.

### 8.3. Generalizability and Future Extensions

The findings reported here are specific to Gaza-related discourse and may not directly transfer to other conflicts, crises, or cultural settings. Expressions of hope and resilience are shaped by local traditions, theology, and collective memory, which vary across societies. Similarly, the focus on informal, dialectal Arabic may limit the applicability of models trained on this dataset to formal or Modern Standard Arabic contexts. Furthermore, the current corpus provides a snapshot in time rather than a longitudinal view of evolving online communication. Future work should expand temporal coverage and include cross-platform and cross-regional sampling to enhance representativeness.

In addition, this study presents the dataset and initial analysis but does not yet include full model development or downstream evaluation. Future research should explore machine-learning applications for automatic detection of hope speech, examine longitudinal changes in online resilience discourse, and integrate multimodal features such as imagery and prosody. These extensions will support more comprehensive and culturally sensitive approaches to modeling constructive communication in Arabic social media. Beyond these extensions, the dataset also provides a foundation for computational modeling of hope-oriented discourse.

**Future Model Development.** While this study focuses primarily on dataset construction and annotation analysis, AraHopeCorpus is designed to support downstream machine learning research on hope speech detection in Arabic. Future work will explore supervised and transformer-based models trained on this corpus, as well as culturally informed fine-tuning strategies for large language models. These experiments will enable a more comprehensive evaluation of automatic hope speech detection and further validate the usefulness of the dataset for Arabic NLP research.

## 9. Ethics Statement

This research adheres to the ethical guidelines for data collection and resource sharing in computational linguistics and digital humanities. All data in *AraHopeCorpus* were collected from publicly available YouTube comments in accordance with the platform's Terms of Service. No private or restricted content was accessed. Usernames, profile images, and other identifiable information were removed to protect individual privacy.

The dataset does not include personally sensitive or demographic information, and all examples were anonymized prior to analysis. Because the material involves conflict-related discourse, care was taken to minimize potential harm by excluding explicit calls for violence or personal attacks. Annotators were trained to approach the content with cultural sensitivity and were informed of the emotional nature of the material. Previous research has highlighted the emotional and psychological challenges faced by annotators working with harmful online content (Al Emadi and Zaghouani, 2024). Given the crisis-related nature of the data in AraHopeCorpus, particular care was taken to support annotators during the annotation process.

The inclusion of ChatGPT as an auxiliary annotator was limited to non-sensitive textual data and followed the principle of human-in-the-loop verification. The dataset will be released for research purposes only under a non-commercial license, accompanied by documentation outlining appropriate and ethical use. This project aligns with the principles of fairness, accountability, and transparency. In addition to procedural compliance, the broader ethical implications and societal relevance of this work are discussed in the section on *Implications and Broader Societal Impact*.

## Acknowledgment

This work was made possible by the National Priorities Research Program grant NPRP14C-0916-210015 from the Qatar National Research Fund (QNRF), part of the Qatar Research, Development and Innovation Council (QRDI).